# Multi-objective Fitness Dependent Optimizer Algorithm

**Jaza M. Abdullah**[1,] , **Tarik A. Rashid** [2 *], **Bestan B. Maaroof** [3], **Seyedali Mirjalili**[4,5]

[1]Information Technology, College of Commerce, University of Sulaimani, Sulaymaniyah, Iraq.
  jaza.abdullah@univsul.edu.iq

[2] Computer Science and Engineering  Department, University of Kurdistan Hewler, Erbil, Iraq.

*Corresponding email address: tarik.ahmed@ukh.edu.krd

[3]Information Technology Department, College of Commerce, University of Sulaimani, Iraq

bestan.maaroof@univsul.edu.iq

[4]Centre for Artificial Intelligence Research and Optimisation, Torrens University, Adelaide, Australia

[5]Yonsei Frontier Lab, Yonsei University, Seoul, Korea

ali.mirjalili@gmail.com

## Abstract

This paper proposes the multi-objective variant of the recently-introduced fitness dependent optimizer (FDO). The algorithm is called a Multi-objective Fitness Dependent Optimizer (MOFDO) and is equipped with all five types of knowledge (situational, normative, topographical, domain, and historical knowledge) as in FDO. MOFDO is tested on two standard benchmarks for the performance-proof purpose: classical ZDT test functions, which is a widespread test suite that takes its name from its authors Zitzler, Deb, and Thiele, and on IEEE Congress of Evolutionary Computation benchmark (CEC-2019) multi-modal multi-objective functions. MOFDO results are compared to the latest variant of multi-objective particle swarm optimization (MOPSO), non-dominated sorting genetic algorithm third improvement (NSGA-III), and multi-objective dragonfly algorithm (MODA). The comparative study shows the superiority of MOFDO in most cases and comparative results in other cases. Moreover, MOFDO is used for optimizing real-world engineering problems (e.g., welded beam design problems). It is observed that the proposed algorithm successfully provides a wide variety of well-distributed feasible solutions, which enable the decision-makers to have more applicable-comfort choices to consider.

## 1. Introduction

Multi-objective optimization problems (MOPs) are in the area of multiple criteria decision-making; the area is also known as multi-objective programming, multi-criteria optimization, or Pareto optimization. Precisely, this area deals with mathematical optimization problems with two or more, often conflicting, objectives.  Research on MOPs became widely popular in 2002 [1]. It is worth mentioning that various real-world applications fall under this area, such as engineering design, energy, economics, logistics, and health science.  Since many real-world problems are MOPs, multi-objective evolutionary algorithms (MOEAs) are used to solve them. Generally speaking, MOEAs include three branches: dominance-based, decomposition-based, and indicator-based evolutionary algorithms (IBEAs) [2] [3] [4].

The first branch of MOEAs includes a non-dominated sorting genetic algorithm (NSGA-II) [5], multi-objective particle swarm optimization (MOPSO) [6], and strength Pareto evolutionary algorithm 2 (SPEA2) [7]. These algorithms are considered posterior optimization algorithms, meaning that they maintain the multi-objective formulation of a multi-objective optimization problem and estimate the Pareto optimal solutions. On the other hand, MOEAs based on decomposition come in the second branch [8]. Such algorithms, use different weights to create different decompositions of the objectives of a multi-objective problem to estimate the Pareto optimal solutions.

The NSGA-II starts with a randomly generated population. As per fitness, it uses a fast-nondominated sorting technique to sort the overall population, and then the children's generation is produced by crossover and mutation. To enhance the diversity level among the solutions, a special crowding distance operator is also applied for later improvement [5]. Moreover, NSGA-II was improved to solve many-objective problems (having four or more objectives) known as NSGA-III [9]. Another popular algorithm with lower computational complexity than NSGA-III is MOPSO [6], which uses an archive grid-based approach to keep diversity among the solutions. In the last decade, many new MOEAs have been reported in the literature; multi-objective cat swarm optimization [10], multi-objective CLONAL algorithm, which is inspired by the clonal selection theory of acquired immunity [11], multi-objective moth flame optimization [12], multi-objective ant lion optimizer [13], multi-objective grey-wolf optimization [14], multi-objective dragonfly algorithm [15], and multi-objective whale algorithm [16]. Having said that, Some MOEAs have been used for production scheduling [17], optimal truss design [18], and resource allocation in cognitive radio networks [19]. Furthermore, some other MOEAs were employed to classify the normal and aggressive behavior of 3D human beings [11].

Finally, the IBEAs have received much popularity due to their strong theoretical support and background [3], IBEAs measure both diversity and convergence of non-dominated solutions in objective space, which are desirable in the context of multi-objective evolutionary optimization [20] [2]. Since IBEAs indicator functions automatically recover the diversity issue among their population solutions, they do not need any diversity maintenance mechanism. The first hypervolume indicator-based EAs known as "hypervolume by slicing objectives (HSO)" was introduced by [21], also local search and hybrid evolutionary algorithms for Pareto optimization were proposed by [22]. As a result, many IBEAs were developed by using different procedures, such as preferences-based information and different local search optimizers, and many others [23] [24] [25] [26]. One obvious disadvantage of IBEA is that it requires additional time for calculating hypervolumes while dealing with many objectives' problems, this issue has been addressed in several types of research, they proposed an enhancement of IBEAs, such as a faster algorithm for calculating hypervolume [27] [28] [29] [30]. To solve MOPs correctly, two factors need to be considered. Firstly, the accuracy of estimated Pareto optimal solutions. Secondly, the diversity of estimated Pareto optimal solutions. Multi-objective meta-heuristics need to address these two, often conflicting, factors. They typically start with a random population of solutions and improve them until the satisfaction of an end condition. The accuracy of the initial population increases over time [31] while some solutions will be favored over others due to them being "close" to the solutions already found in the objective space.

Many efforts were made regarding the local and global guide selection, such as using the adaptive grid to select the global guide and introducing an extra repository to store the non-dominated particles in MOPSO [6]. In another attempt, the global focus is selected using crowding distance in crowding distance MOPSO [32]. In [33], Mostaghim and Teich determined the local guide based on the sigma method, while Pulido and Coello [34] used a clustering technique for the same purpose. On the other hand, a genetic operator and special domination principle have been employed in terms of population diversity to improve the Pareto front variety. Zitzler proposed the elitism mechanism that uses crossover and mutation on individuals that have been selected from the combination of population and repository [35], while Laumanns proposed Ɛ-box dominance to combine diversity and convergence [36]. In [37] and then [31] a simulated binary crossover (SBX) is used. Additionally, in Pareto entropy MOPSO, a cell distance-based individual density is used to select the global guide [38]. Moreover, a hybrid grey wolf optimizer uses to increase the efficiency of complex industrial system designs [39]. Furthermore, [40] developed a modified genetic algorithm (MGA), and they used it for the ship routing and scheduling model. Another modification of GA has been used by [41] for utilizing a reduced-order model for preliminary reactor design optimization. Recently, a new multi-objective learner performance-based behavior algorithm (MOLPB) was used to solve a four-bar truss design problem, pressure vessel design, coil compression spring design problem, speed reducer design problem, and car side impact design problem [42].

According to [43], a cultural algorithm is formed of five types of knowledge: situational, normative, topographical, domain, and historical knowledge, These file types are explained briefly in the following list:

1. Situational knowledge is a set of objects useful for the experience interpretation of all individuals in a certain population. In other words, situational knowledge guides individuals to move toward exemplars (best local or best global guides).
2. Normative knowledge: includes a set of promising ranges of decision variables. It offers strategies for individual adjustments. More precisely, it leads individuals to dive into a good range.
3. Topographical knowledge: it splits the completely feasible search landscape into cells. Each of these cells represents a different spatial characteristic; also, each cell selects the best individual in its specific ranges. Keeping the idea simple, topographical knowledge leads individuals toward the best cell.
4. Domain knowledge: it records information about the problem domain to guide the whole search; it is considered useful during the search process.
5. Historical knowledge: it records the key events in the search process by keeping track of significant individuals' history. Key events might be a big move in the search space or sometimes comes in the form of notable changes in the search landscape. Individuals are using historical knowledge to select a preferable direction.

Various research has been conducted in the field of nature-inspired metaheuristic algorithms; additionally, many efficient algorithms have been proposed in the literature. Alternatively, there is always room for new algorithms, as long as the proposed algorithm provides better or comparative performances, as explained by [44] in their work titled "No Free Lunch Theorems for Optimization" in 1997. Thus, there is no single global algorithm that can provide the optimum solution for every optimization problem. Furthermore, this work represents a multi-objective mode of the currently existing single-objective algorithm called FDO [45]. One major limitation of many MOEAs is that they tend to fall into local optimum in high-dimensional space easily and have a low convergence rate in the iterative process [46], in MOFDO, the fitness weight and weight factor parameters were used to increase both coverage and convergence of the algorithm (more on this is discussed in section 2.2), also storing previous good decision for later reuse will convergence speed as well. For these reasons, a new algorithm called MOFDO is proposed in this work. This algorithm is inspired by the swarming behavior of bees during the reproductive process when they search for new hives. The proposed algorithms have nothing in common with the artificial bee colony (ABC) algorithm (except both algorithms are inspired by bee behavior, and both are nature-inspired meta-heuristic algorithms).

Regarding this paper's major contributions, a new novel multi-objective mode of the novel single-objective FDO algorithm is proposed. One of the major contributions of this work is developing MOEA, which has a linear time and space complexity (more on this is discussed in section 2.4). Moreover, besides having an archive for saving the Pareto front solutions, a polynomial mutation mechanism is employed as a variation operator. Furthermore, extra storage has been used for saving the previous paces for the potential reuse in the next iterations; this will improve the algorithm performance Additionally, Hypercube grids are used in the implementation to help select the local and global guide individuals.

The rest of this paper is organized as follows: Section 2, explains the methodology of theoretical calculations, also the definitions for the Pareto sets, Pareto optimality set, and Pareto front set. Moreover, the MOFDO is proposed and mathematically explained in detail. Then in Section 3, results and discussion are discussed. In Section 4, MOFDO is employed to solve a real-world engineering problem (welded beam design problem). Finally, the conclusions are outlined in Section 5.

## 2. Methodology

In this section, some of the preliminaries and essential definitions of multi-objective optimization are covered. MOFDO algorithm is mathematically and programmatically presented in detail. The level of detail is constructed in a way that other researchers can easily replicate our work.

### 2.1. Pareto Optimal Solutions Set

Mathematically speaking, MOPs can be represented as follows, with no loss of generality:

minimize: $\vec{F}(\vec{x}) = \{f_1(\vec{x}), f_2(\vec{x}), \dots, f_n(\vec{x})\}$ (1)

subject to:

$$g_{i(\vec{x})} \leq 0, \quad i = 1, 2, \dots, m$$

$$h_{i(\vec{x})} = 0, \quad i = 1, 2, \dots, p$$

where: $n$ is several objectives, $g$ and $h$ are constraints, $m$ is inequality constraint and $p$ is equality constraint [47].

This type of problem cannot be optimized normally with the traditional single-objective algorithm, not just because of its multi-objective nature but also because of conflicting objectives in the same optimization problem, which means there is no single optimum solution. Instead, there is a set of optimal solutions known as the Pareto optimality solutions set, representing the best trad-offs between objectives. For readability purposes, Pareto optimality will be discussed briefly in this sub-section. Pareto optimality solutions can be explained using the following definitions [48]:

- Def. #1: for vectors (solution) $\vec{a}$ and $\vec{b}$ in optimization problem $K^t$. For $i = 1, 2, \dots, m$, $\vec{a} \leq \vec{b}$ if the objectives of vector $\vec{a}$ smaller or equal to the objectives of vector $\vec{b}$ and at least there is $\vec{a_i} < \vec{b_i}$.
- Def. #2: if $\vec{a} \leq \vec{b}$ then: $\vec{a}$ dominates $\vec{b}$, and denoted by $\vec{a} \prec \vec{b}$.
- Def. #3: two solutions might not dominate each other if Def. #1 is not applied, in this case, solutions $\vec{a}$ and $\vec{b}$ are non-dominated concerning each other, and denoted as $\vec{a} \not\prec \vec{b}$ is the set of all nondominated known as the Pareto optimal solution set $P_s$, and defined as equation (2):

$$P_s := \{\vec{a}, b \in A | \exists F(a) \succ F(b)\} \tag{2}$$

- Def #4: The set holding equivalent object values of Pareto optimal solutions in $P_s$, is known as Pareto optimal front $P_f$, and defined as equation (3):

$$P_f := \{F(\vec{a}) | \vec{a} \in |P_s\} \tag{3}$$

**2.2. Multi-Objective Fitness Dependent Optimizer**

Multi-objective fitness dependent optimizer (MOFDO) is based on our recent work, a single objective fitness dependent optimizer FDO [45]. FDO is a metaheuristic algorithm, the bee swarming reproductive process, and their collective decision-making has inspired this algorithm. FDO updates the individual position by adding $pace$ to the current position as shown in equation (4); the same mechanism is also applied in MOFDO. However, to calculate the $pace$, the conditions presented in Equations (5, 6, and 7) need to be considered, and these conditions depend on the fitness weight ($fw$) value. $fw$ can be calculated using the problem cost function values according to Equation (8). It is worth mentioning that the $pace$ represents both domain and historical knowledge in MOFDO.

$$X_{i,t+1} = X_{i,t} + pace \tag{4}$$

where $i$ represents the current individual number, $t$ represents the current iteration, $x$ is the individual itself, and $pace$ is the movement rate and direction. Recalling that the $pace$ value mostly relies on the $fw$. However, the direction of $pace$ (value sign) entirely depends on a random mechanism.

$$\begin{cases} fw = 1 \text{ or } fw = 0 \text{ or } \sum_{o=1}^{n} x_{i,t\ fitnees_o} = 0, \quad pace = x_{i,t} * r & (5) \\ fw > 0 \text{ and } fw < 1 \begin{cases} r < 0, pace = (x_{i,t} - x_{i,t}^*) * fw * (-1) & (6) \\ r \geq 0, \quad pace = (x_{i,t} - x_{i,t}^*) * fw & (7) \end{cases} \end{cases}$$

Equations (5, 6, and 7) contain two different conditions. Firstly, if $fw$ is equal to zero, or, if $fw$ is equal to one, or if $\sum_{o=1}^{n} x_{i,t\ fitnees_o} = 0$, then the $pace$ should be calculated as Equation (5). Secondly, if $fw$ value comes in between zero and one, then the random number $r$ is generated in the [-1, 1] range, if $r$ is a negative number, then Equation (6) will be used; otherwise, Equation (7) will be used for calculating the $pace$. $fw$ Can be computed using Equation (8):

$$fw = \left| \frac{\sum_{o=1}^{n} x_{i,t\ fitness_o}^*}{\sum_{o=1}^{n} x_{i,t\ fitnees_o}} \right| - wf \tag{8}$$

where $\sum_{o=1}^{n} x_{i,t\ fitness_o}^*$ is a sum of the cost function of the global best individual, $n$ is the number of objectives, and $o = \{1, 2, \dots, n\}$, the $\sum_{o=1}^{n} x_{i,t\ fitnees_o}$ is the sum of the current individual's cost function, again $n$ is the number of objectives, and $o = \{1, 2, \dots, n\}$. Finally, $wf$ in Equation (8) is a weight factor, and its value is either 0 or 1. One may notice that, when $wf = 0$, it does not affect the equation and can be ignored. Interested readers are referred to [45] for more details about single-objective FDO.

Although the algorithm structure is the same as a single objective FDO to some extent, there are several additional improvements in MOFDO as follows:

1- An archive (repository) is used for holding Pareto front solutions during optimization, as it has been widely used in the literature for this purpose [5].

2- Before adding the nondominated solution to the archive, a polynomial mutation is applied to Pareto front solutions. The polynomial mutation has been employed in MOEAs as a variation operator [49], and it is defined as Equation (9) [50]:

$$S_i = (x_1, x_2, \ldots, x_n)$$
$$S_{Ni}(x_j) = S_i(x_j) + \alpha \cdot \beta_{max}(x_j), \quad i = 1,2,\ldots,NP, \quad j = 1,2,\ldots n$$
$$\alpha = \begin{cases} (2v)^{\frac{1}{(q+1)}} - 1, & v < 0.5 \\ 1 - (2(1-v))^{\frac{1}{(q+1)}}, & otherwise \end{cases} \quad (9)$$
$$\beta_{max}(x_j) = Max[S_i(x_j) - l_j, u_j - S_i(x_j)], i = 1,2,\ldots,NP, j = 1,2,\ldots n$$

where $S_{Ni}$ is a new solution, $S_i(x_j)$ is a current solution, $\beta_{max}(x_j)$ is the maximum perturbation acceptable between the original and mutated solution, *NP is the* population size, *q* is a positive real number, $v$ is a uniformly distributed random number in the [0, 1] range, *l* is a lower boundary of the decision variable *x*, u is an upper boundary of decision variable *x*, and *n* number of decision variables (problem dimensions).

3- In MOPs, the fittest solution cannot simply be chosen as a global guide (normative knowledge), as this might be the case in single-objective optimization since there is more than one objective. Usually, these objectives conflict with each other. Therefore, selecting a global guide needs a more careful decision. In this work, the global guide individual is denoted by $x^*$, a global-best nondominated solution selected from the least populated region by artificial scout bees the same as to [51] work on MOPSO. For this purpose, a mechanism called archive controller is used to divide the archive into multiple equally sized grids (sub-hyper-spheres in multi-dimension problems) [15], in this work, these have been called hypercube grids, which represent a topographical knowledge usage in MOFDO. The hypercube grid mechanism allows the algorithm to determine the least populated area simply by counting the number of solutions in each grid. The global best solution will be selected from the least populated area, see Figure (1).

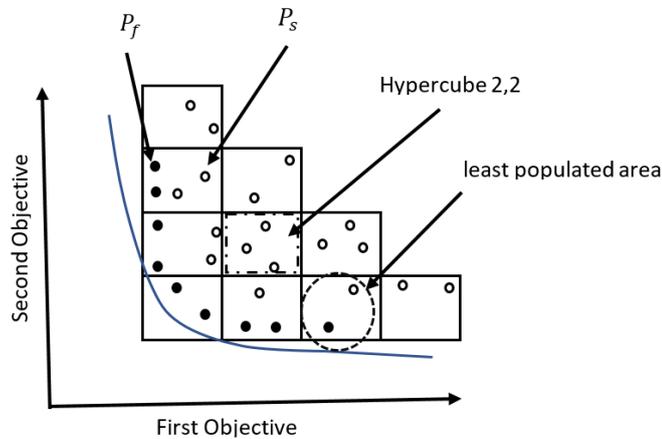

Figure 1. shows Pareto solution, Pareto front, and hypercube grids which is provide helps in selecting global and local guide

The reason behind selecting the global guide from the least populated area is to maintain a good diversity in the obtained Pareto front solutions. As a result, decision-makers will have more diverse choices (solutions) to consider. Nevertheless, the archive has a limited size. When a new non-dominated solution is found, and the archive is already reached its maximum capacity, the archive controller removes the worst solution from the most populated grid. Hence, the newly discovered solution can fit in, as long as the new solution is better than the archive's worst solution.

4- Regarding selecting the personal guide (situation knowledge), the same hypercube grid mechanism has been used for dividing the search landscape into equally sized cells, then inside each cell, the best personal solution is selected as a local guide.

## 2.3. Multi-Objective Fitness Dependent Optimizer Working Mechanisms

The MOFDO starts by randomly distributing the search individuals over the search space as presented in the pseudocode Figure (2), and more explanations are given in the flowchart Figure (3). Then, an archive with a specific size is created, and hypercube grids are generated. From here, the main algorithm loop will start in Line (4), which mainly depends on a specific number of iterations or until a certain condition is met. In line (5), for each search, individual (artificial scout bee) operations from Lines (6 to 27) will be repeated according to the number of individuals. The mentioned operations include: finding the global best search agent, finding $fw$ using equation (8), Lines (10 to 12) applying conditions from Equations (5, 6, and 7) to calculate $pace$ then afterward, Line (13) calculating a new search agent position using equation (4). When the new search agent is discovered, the algorithm always checks whether the new result (cost function) dominates the old result or not (14). If it is, then the new position will be accepted, and the $pace$ will be stored for potential reuse in the future, as shown in Line (15). However, if it is not, if previously saved $pace$ available, it will be used instead of the new one, hoping to generate a better result unless the search agent maintains the current position (see Lines (17 to 22). The polynomial mutation will apply to get more variant solutions in Line (24) and then check whether the solution can fit inside the archive or not in Lines (25 and 26). Hypercube grid indices are always updated according to the search landscape changes in Line (27).

```
1   Initialize the artificial scout bee population $X_{i,t}$ $i = 1, 2, ..., n$, and $t=1,2, ..., m$
2   Creating archive for nondominated solutions
3   Generate Hypercube grid
4   While (t) iteration limit not reached (m) (or solution good enough)
5       for each artificial scout bee $X_{i,t}$
6           find best artificial scout bee $X_{i,t}^*$
7           generating random walk r in [-1, 1] rang
8           calculating fitness weight value. Equation ( 7 )
9           //checking Equation ( 4 , 5, and 6 ) conditions
10          if (fitness_weight ≥ 1 or fitness_weight ≤ -1 or $\sum_{o=1}^{n} x_{i,t\ fitnees_o} = 0$)
11              fitness_weight = r
12          end if
13          calculate $X_{i,t+1}$    equation (3 )
14          if( $X_{i,t+1}$ fitnesses dominate $X_{i,t}$ fitnesses)
15              move accepted and pace  saved
16          else
17              calculate $X_{i,t+1}$ equation (3 )  with previous pace
18              if ($X_{i,t+1}$ fitnesses dominate $X_{i,t}$ fitnesses)
19                  move accepted and pace  saved
20              else
21                  maintain current position (don't move)
22              end if
23          end if
24          Apply polynomial mutation
25          Add non-dominated Bees (solutions) to archive
26          Keep only non-dominated members in the archive
27          Update Hypercube Grid Indices
28      end for
29   end while
```

Figure 2.  MOFDO Pseudocode

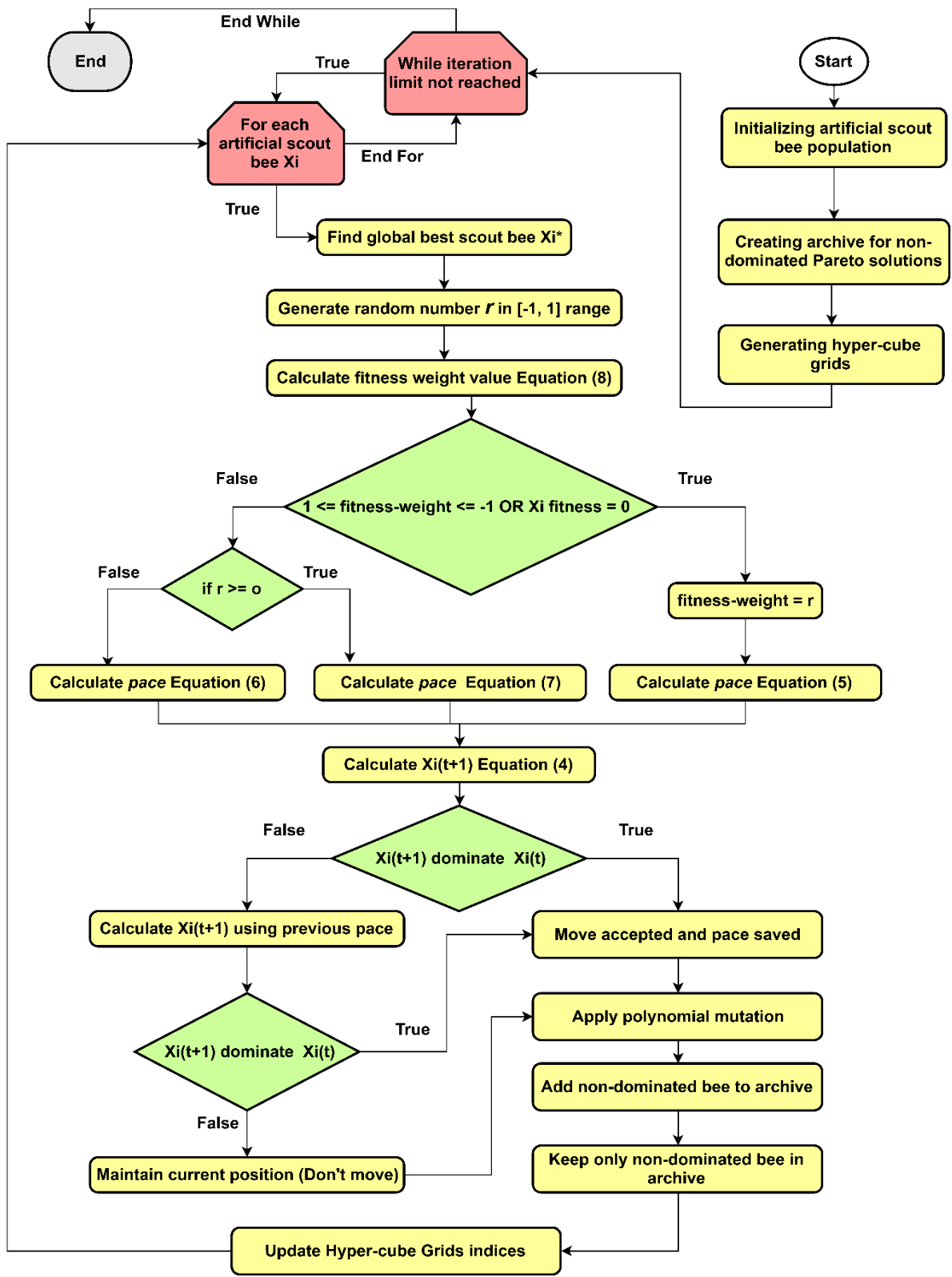

Figure 3. flowchart shows how MOFDO works programmatically

## 2.4. Multi-Objective Fitness Dependent Optimizer Algorithm Time and Space Complexity

Generally, computational complexity is mainly concerned with the time and space required to solve a given problem, Regarding MOFDO mathematical complexity, it has an O(*p*n* + *p*CF*) linear time complexity in each iteration, where *p* is the population size, *n* is the dimension of the problem, and *CF* is the cost of the objective function. Whereas, it has an O(*p*CF* +*p*pace*) space complexity for all iterations, where the *pace* is the best previous paces stored. Thus, MOFDO time complexity is proportional to the number of iterations. However, its space complexity will be the same during iterations.

Nonetheless, MOFDO has a simple objective value calculations calculation mechanism, it has only (a random number and fitness weight) to be calculated for each agent, whereas, in MOPSO for calculating each solution, there are global best, agent best, and search factors C1 and C2, and random numbers (R1 and R2 parameters) to be calculated [33]. Also, in the MODA, there are five different parameter weights to be calculated (attraction, distraction, separation, alignment, cohesion, and some random values), and most of these parameters have accumulative nature (summation and multiplication), and their values depend on all other agents' value, resulting in even more complex calculations [14]. Finally, according to Currya and Daglia, NSGA-III has a mathematical complexity of O($n_g * n_o * n_p^2$), where $n_o$ is the number of objectives, $n_p$ is the population size, and $n_g$ is the number of generations, $n_g$ can have any complexity from constant to $n_P$ depending upon the stopping criteria used [52]. From here, it can be seen that NSGA-III has an order of $n^2$ complexity, which is more complex than MOFD linear complexity.

## 3. Results and Discussion

For testing MOFDO algorithm performance, two different types of multi-objective test functions were selected: Classical ZDT benchmarks [35] and 2019 CEC Multi-modal multi-objective benchmarks [53]. The MOFDO results are compared to the results of the latest state of the art of MOPSO, NSGA-III, and modern multi-objective dragonfly algorithm (MODA) [15].

### 3.1. Classical ZDT Benchmark Results.

This benchmark includes five well-known challenging test functions from ZDT1 to ZDT5. Their mathematical definition is presented in Table (7) (See Appendix), the MOFDO results are compared to three well-known algorithms: MOPSO, MODA, and NSGA-III. The results are shown in Table (1), each algorithm is allowed to run for 500 iterations, each equipped with initial 100 search individuals and an archive size of 100, the parameter settings for each algorithm are as described in their original papers [15] [53] [5]. However, the parameter settings for MOFDO are:

*Polynomial Mutation Rate = 5.*
*Number of Grids per Dimension = 7.*
*Best Bee Selection Factor = 2.*
*Delete Factor =2.*
*Inflation Rate =1.*

The inverse generational distance (IGD) as shown in equation (10), is a measurement, which uses a true Pareto front of the problem as a reference, then compares each of its elements concerning the $P_f$ produced by the algorithm as described by [54].

$$IGD = \frac{\sqrt{\sum_{i=1}^{n} d_i^2}}{n} \qquad (10)$$

where $d_i$ is the Euclidean distance between the closest obtained Pareto optimal solutions and the $i^{th}$ true Pareto optimal solution in the reference set, and $n$ is the number of true Pareto optimal solutions, it should be clear that a value of IGD = 0 indicates that all the elements generated are in the true Pareto front of the problem. The IGD results of 30 independent runs are collected for each algorithm, then the average (mean), standard deviation STD, best, and worst IGD are calculated, see Table (1).

TABLE (1)
Classical ZDT Benchmark results

| Functions | Algorithms | IGD AVG. | IGD STD. | IGD Best | IGD Worst |
|---|---|---|---|---|---|
| ZDT 1 | **MOFDO** | **0.06758** | **0.030911** | **0.0018** | **2.61533** |
|  | MODA | 0.07653 | 0.012071 | 0.0420 | 0.59398 |
|  | MOPSO | 0.07843 | 0.008848 | 0.0446 | 1.14508 |
|  | NSGA-III | 0.52599 | 0.509184 | 0.0134 | 3.69236 |
| ZDT 2 | MOFDO | 0.03511 | 0.00404 | 0.0207 | 0.0515 |
|  | **MODA** | **0.00292** | **0.00026** | **0.0002** | **0.0116** |
|  | MOPSO | 0.03243 | 0.00093 | 0.0212 | 0.0682 |
|  | NSGA-III | 0.13972 | 0.02626 | 0.1148 | 0.1834 |
| ZDT 3 | **MOFDO** | **0.06676** | **0.023913** | **0.0014** | **2.2206** |
|  | MODA | 0.07653 | 0.014411 | 0.0401 | 0.8267 |
|  | MOPSO | 0.07758 | 0.005755 | 0.0427 | 1.0355 |
|  | NSGA-III | 0.19474 | 0.080043 | 0.1935 | 0.1962 |
| ZDT4 | MOFDO | 0.68020 | 0.352945 | 0.2679 | 1.6776 |
|  | MODA | 64.9628 | 2.847807 | 51.742 | 500.93 |
|  | **MOPSO** | **0.46175** | **0.047785** | **0.2515** | **5.1602** |
|  | NSGA-III | 0.73731 | 0.307518 | 0.7360 | 0.7387 |
| ZDT5 | MOFDO | 0.35853 | 0.161795 | 0.1221 | 1.9125 |
|  | MODA | 0.11349 | 0.018270 | 0.0142 | 2.3938 |
|  | **MOPSO** | **0.26862** | **0.136598** | **0.0468** | **2.1894** |
|  | NSGA-III | 0.66397 | 0.235754 | 0.66273 | 0.66545 |

TABLE (2)
The ranking table shows algorithms performances in Table (1)

| Functions | MOFDO Rank | MODA Rank | MOPSO Rank | NSGA-III Rank |
|---|---|---|---|---|
| ZDT1 | 1 | 2 | 3 | 4 |
| ZDT2 | 3 | 1 | 2 | 4 |
| ZDT3 | 1 | 2 | 3 | 4 |
| ZDT4 | 2 | 4 | 1 | 3 |
| ZDT5 | 3 | 1 | 2 | 4 |
| Total Ranking | **10** | **10** | **11** | **19** |

* Pareto front
o Pareto solution

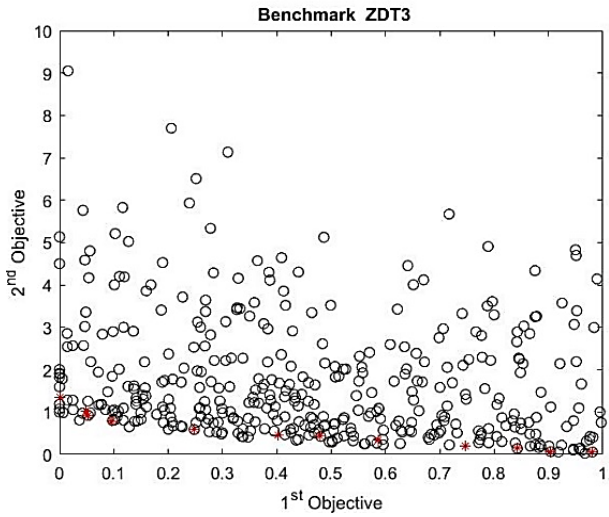

(a) MOFDO in iteration 4, found 23 $P_{fs}$

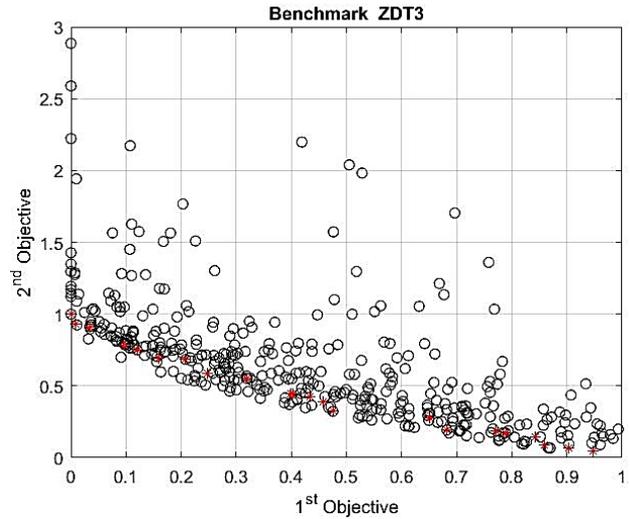

(b) MOFDO in iteration 19, found 90 $P_{fs}$

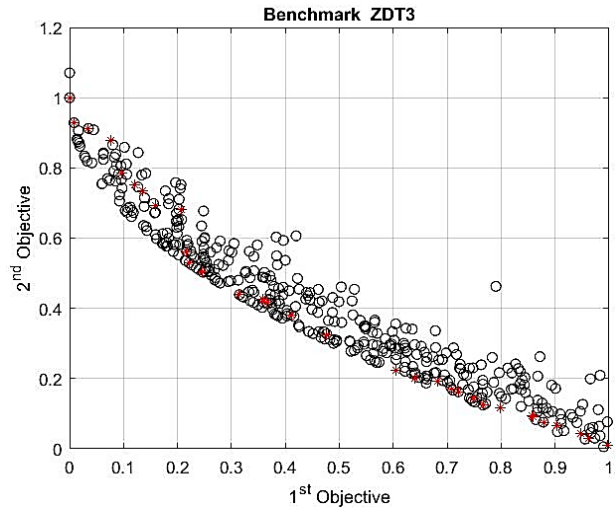

(c) MOFDO iteration in 93, found 174 $P_{fs}$

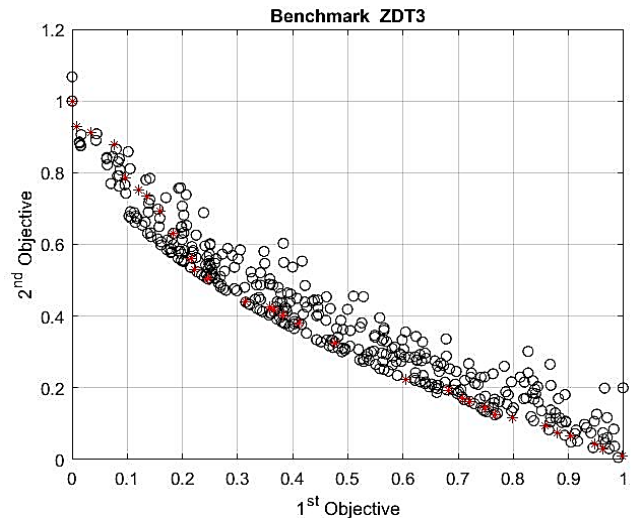

(d) MOFDO in iteration 200, found 400 $P_{fs}$

Figure 4. show how MOFDO solved the ZDT3 test function from an initially random solution toward Pareto front optimality.

Here ranking tables are used to show the rank of each algorithm, see Tables (2 and 4), for example, MOFDO came at the first position in ZDT1, then its rank is equal to 1, MOFDO came at third position in ZDT2, then its rank is equal to 3 and so on, the total rank represents the summation of all acquired ranks by certain algorithm, the ranking table is a simple way to show the superiority of certain algorithm among the group of competed algorithms. Also, the ranking tables were used in the Friedman test for all test functions, which shows whether the results are statistically significant or not (see Section 3.3).

As can be seen in the ranking-table Table (2), MOFDO outperforms MOPSO and NSGA-III in most cases; yet, it provides comparative results compared to MODA. MOFDO and MODA achieve a total ranking of 10. Figure (4) shows the ZDT3 solution landscape as an example of MOFDO performance, Figure 4(a) shows how initially randomly distributed solutions which only 23 $P_{fs}$ is located, then throughout iterations, MOFDO successfully increased the number of well-distributed $P_f$ solutions as shown in Figures 4(b, c, and d).

### 3.2. CEC 2019 Multi-Modal Multi-Objective Benchmarks

A set of 12 CEC-2019 multi-modal multi-Objective (MMO) 2019 benchmarks are selected as described by [55], and their mathematical definition is shown in Table (8) (See Appendix). The reason behind selecting this benchmark is, that these test functions are represented a more difficult challenge than the ZDT benchmark for MOFDO; they have different characteristics, such as problems with different shapes of PSs and PFs, with the coexistence of local and global PSs, also having a scalable number of PSs, decision variables, and objectives. MOFDO results compared to MOPSO, MODA, and NSGA-III as shown in Table (3). The results are explained by the ranking system in Table (4), which shows the MOFDO ranked in first place with superior results in most cases; MODA comes in second place, then MOPSO and NSGA-III.

TABLE (3)
CEC 2019 MMF Benchmark results

| Functions | Algorithms | IGD AVG. | IGD STD. | IGD Best | IGD Worst |
|---|---|---|---|---|---|
| MMF1 | MOFDO | 0.18401921 | 0.0454458 | 0.0882267 | 2.2406685 |
| | MODA | 0.87703300 | 0.5302916 | 0.3618665 | 9.6322659 |
| | MOPSO | 0.32173518 | 0.1108645 | 0.1419164 | 0.5425832 |
| | **NSGA-III** | **0.00351527** | **0.0005796** | **0.0022760** | **0.0049877** |
| MMF2 | **MOFDO** | **0.09108902** | **0.0237087** | **0.0377645** | **0.9732300** |
| | MODA | 0.41152959 | 0.3041183 | 0.0883137 | 16.525584 |
| | MOPSO | 0.27264430 | 0.0606373 | 0.1906970 | 1.1093294 |
| | NSGA-III | N/A | N/A | N/A | N/A |
| MMF3 | **MOFDO** | **0.09121177** | **0.0184429** | **0.0412612** | **1.0219979** |
| | MODA | 0.38999723 | 0.3195349 | 0.0720422 | 20.350064 |
| | MOPSO | 0.46594619 | 0.1566197 | 0.3377674 | 1.7063454 |
| | NSGA-III | 6.22408148 | 2.4773195 | 0.0523665 | 17.564674 |
| MMF4 | MOFDO | 0.08195533 | 0.0340485 | 0.0453016 | 0.1879044 |
| | **MODA** | **0.00781723** | **0.0038766** | **0.0003086** | **0.0319178** |
| | MOPSO | 0.06806595 | 0.0056268 | 0.0437656 | 0.1621893 |
| | NSGA-III | 0.04784677 | 0.0102452 | 0.0054074 | 0.1314727 |
| MMF5 | **MOFDO** | **0.08166825** | **0.0203041** | **0.0410373** | **0.2872334** |
| | MODA | 0.20697935 | 0.1068910 | 0.1177007 | 0.2665746 |
| | MOPSO | 0.36516458 | 0.0818221 | 0.1431733 | 0.6443415 |
| | NSGA-III | 0.23876644 | 0.1195411 | 0.2370972 | 0.2405229 |
| MMF6 | **MOFDO** | **0.06319825** | **0.0052717** | **0.0435369** | **0.3023473** |
| | MODA | 6.20722767 | 7.0529532 | 3.8844812 | 18.481523 |
| | MOPSO | 0.57534658 | 0.2975283 | 0.2154710 | 1.0872217 |
| | NSGA-III | 0.70090140 | 0.2650748 | 0.6990628 | 0.7024793 |
| MMF7 | **MOFDO** | **0.14853951** | **0.0219769** | **0.0870897** | **0.2362694** |
| | MODA | 0.36139133 | 0.1036987 | 0.1322042 | 1.0331912 |
| | MOPSO | 0.33104321 | 0.0493751 | 0.1186243 | 0.4153237 |
| | NSGA-III | 0.40264339 | 0.1635305 | 0.4014061 | 0.404124 |
| MMF8 | MOFDO | 0.15550447 | 0.0869989 | 0.0343290 | 1.8046810 |
| | MODA | 0.08058735 | 0.2652865 | 0.0056401 | 10.189273 |
| | MOPSO | 0.14156195 | 0.1003047 | 0.0450518 | 2.6718013 |
| | **NSGA-III** | **0.01038634** | **0.0032172** | **0.0041821** | **0.0276770** |
| MMF9 | MOFDO | 0.47321267 | 0.1219659 | 0.3404164 | 1.7427477 |
| | **MODA** | **0.05060995** | **0.0274896** | **0.0051946** | **0.3219936** |
| | MOPSO | 1.33275589 | 0.1427530 | 0.7792982 | 2.0541908 |
| | NSGA-III | 0.96369603 | 0.3011030 | 0.0027674 | 0.2438216 |
| MMF10 | MOFDO | 0.44207841 | 0.1277887 | 0.3104489 | 1.1022388 |
| | **MODA** | **0.09017605** | **0.0385574** | **0.0039308** | **0.3893014** |
| | MOPSO | 1.00054897 | 0.1542964 | 0.7005662 | 1.4956293 |
| | NSGA-III | 3.89641261 | 4.6634273 | 0.0028740 | 4.4067242 |
| MMF11 | **MOFDO** | **0.09260275** | **0.0209854** | **0.0635536** | **0.2692325** |
| | MODA | 0.09291338 | 0.0515551 | 0.0042148 | 0.4962967 |
| | MOPSO | 1.30789085 | 0.1622864 | 0.6847497 | 2.2372910 |
| | NSGA-III | 1.18058557 | 0.7034533 | 0.0034136 | 4.2312541 |
| MMF12 | MOFDO | 0.08314653 | 0.0217281 | 0.0554114 | 0.2901663 |
| | **MODA** | **0.03661122** | **0.0119014** | **0.0035018** | **0.1841556** |
| | MOPSO | 0.13651933 | 0.0237385 | 0.0667090 | 0.3416214 |
| | NSGA-III | 0.35064339 | 0.1613096 | 0.3494061 | 0.352124 |

TABLE (4)
THE RANKING TABLE SHOWS ALGORITHMS PERFORMANCES IN TABLE (3)

| Functions | MOFDO Rank | MODA Rank | MOPSO Rank | NSGA-III Rank |
|---|---|---|---|---|
| MMF1 | 2 | 4 | 3 | 1 |
| MMF2 | 1 | 3 | 2 | 4 |
| MMF3 | 1 | 2 | 3 | 4 |
| MMF4 | 4 | 1 | 2 | 3 |
| MMF5 | 1 | 2 | 3 | 4 |
| MMF6 | 1 | 4 | 3 | 2 |
| MMF7 | 1 | 2 | 3 | 4 |
| MMF8 | 4 | 2 | 3 | 1 |
| MMF9 | 2 | 1 | 4 | 3 |
| MMF10 | 2 | 1 | 3 | 4 |
| MMF11 | 1 | 2 | 4 | 3 |
| MMF12 | 2 | 1 | 3 | 4 |
| **TOTAL RANKING** | **22** | **25** | **36** | **37** |

To prove whether the results of Tables (1 and 3) are statistically significant or not, the Wilcoxon rank-sum test has been conducted to find the p-value between the MOFDO and other algorithms. As presented in Table (5), the majority of the results in Table (1) (ZDT benchmarks results) are statistically significant since the p-value is smaller than 0.05.

TABLE (5)
THE WILCOXON RANK-SUM TEST (P-VALUE) FOR ZDT BENCHMARKS

| Functions | MOFDO Vs. MOPSO | MOFDO Vs. MODA | MOFDO Vs. NSGA-III |
|---|---|---|---|
| ZDT1 | 0.069585315 | 0.144884804 | 1.06115E-05 |
| ZDT3 | 0.000828462 | 5.73525E-46 | 0.002956318 |
| ZDT3 | 0.01911876 | 0.06013187 | 1.35925E-11 |
| ZDT4 | 0.001385852 | 9.07401E-72 | 0.506664868 |
| ZDT5 | 0.023547289 | 2.40076E-11 | 2.39778E-07 |

Nonetheless, Table (6) also shows the level of significance of the results in Table (3) (CEC 2019 benchmarks results), as clearly it can be seen, that the results are statistically significant in almost all cases, except three cases in MMF8 and MMF9.

TABLE (6)
THE WILCOXON RANK-SUM TEST (P-VALUE) FOR CEC 2019 BENCHMARKS

| Functions | MOFDO Vs. MOPSO | MOFDO Vs. MODA | MOFDO Vs. NSGA-III |
|---|---|---|---|
| MMF1 | 4.41881E-08 | 1.76049E-09 | 1.41173E-29 |
| MMF2 | 5.48551E-22 | 3.45868E-07 | N/A |
| MMF3 | 7.46013E-19 | 3.72574E-06 | 1.23675E-19 |
| MMF4 | 0.031476101 | 3.98328E-17 | 2.22201E-06 |
| MMF5 | 6.57481E-26 | 4.20625E-08 | 2.01881E-09 |
| MMF6 | 2.64131E-13 | 1.27723E-05 | 4.39284E-19 |
| MMF7 | 5.34254E-26 | 8.11656E-16 | 1.14766E-11 |
| MMF8 | 0.567417814 | 0.147031583 | 3.58707E-12 |
| MMF9 | 8.12705E-33 | 5.09201E-26 | 5.83029E-07 |
| MMF10 | 5.57562E-22 | 7.29596E-21 | 5.74382E-06 |
| MMF11 | 2.44568E-44 | 0.975720942 | 1.36419E-10 |
| MMF12 | 9.63988E-13 | 1.07115E-14 | 1.31857E-12 |

### 3.3. Friedman Test for the Results

Friedman test has been used to show that the given results are statically significant, the Friedman test analyzes whether there are statistically significant differences between three or more dependent samples, it is the non-parametric counterpart of the analysis of variance with repeated measures [56], Friedman test has been used to show that the previous results in Tables (1 and 3) are statistically significant.

In the Friedman test, there are two different hypotheses to accept, the null hypothesis is, that there are no significant differences between the dependent groups, and the alternative hypothesis is, that there is a significant difference between the dependent groups, the Friedman test does not use the true values but the ranks of the values. Friedman test can be calculated as Equation (11).

$$x_r^2 = \frac{12}{nk(k+1)} \sum R^2 - 3n(k+1) \tag{11}$$

Where $x_r^2$ is a Chi-square, $n$ is several test functions, $k$ represents the number of groups (number of compared algorithms), and $R$ is the square root of the total rank of each group, using the total rank from Tables (2 and 4) From here, to find the state decision rule, the degree of freedom is needed, which can be found as $df = k -1= 4-1= 3$, according to Chi-square distribution table of alpha significance level, the decision rule state of df =3 is equal to 7.815 for a *p-value* < 0.5 [57].

The Friedman test calculation for Tables (2 and 4) together would be:

Total Rank of *R: MOFDO =32 , MODA = 35, MOPSO = 47* and *NSGA-III = 56*

$n= 17$ since there are 17 test functions results in both Tables (2 and 4), and $k = 4$

$$x_r^2 = \frac{12}{17*4(4+1)} \sum (32^2 + 35^2 + 47^2 + 56^2) - 3*17(4+1)$$

$x_r^2 = 13.0235$ and the p-value is .00459.

According to the Friedman test, the null hypothesis is rejected, since the result is greater than the 7.815 Chi-square distribution, which means the results are statistically significant at a p-value < 0.5.

Finally, as a performance proof, Figure (5) shows the MMF4 benchmark solution landscape as an example, Figure 5(a) shows the initial random distribution of only 31 $P_{fs}$ solutions, then throughout iterations, the number of well-distributed $P_f$ solutions are increased in Figures 5(b, c, and d).

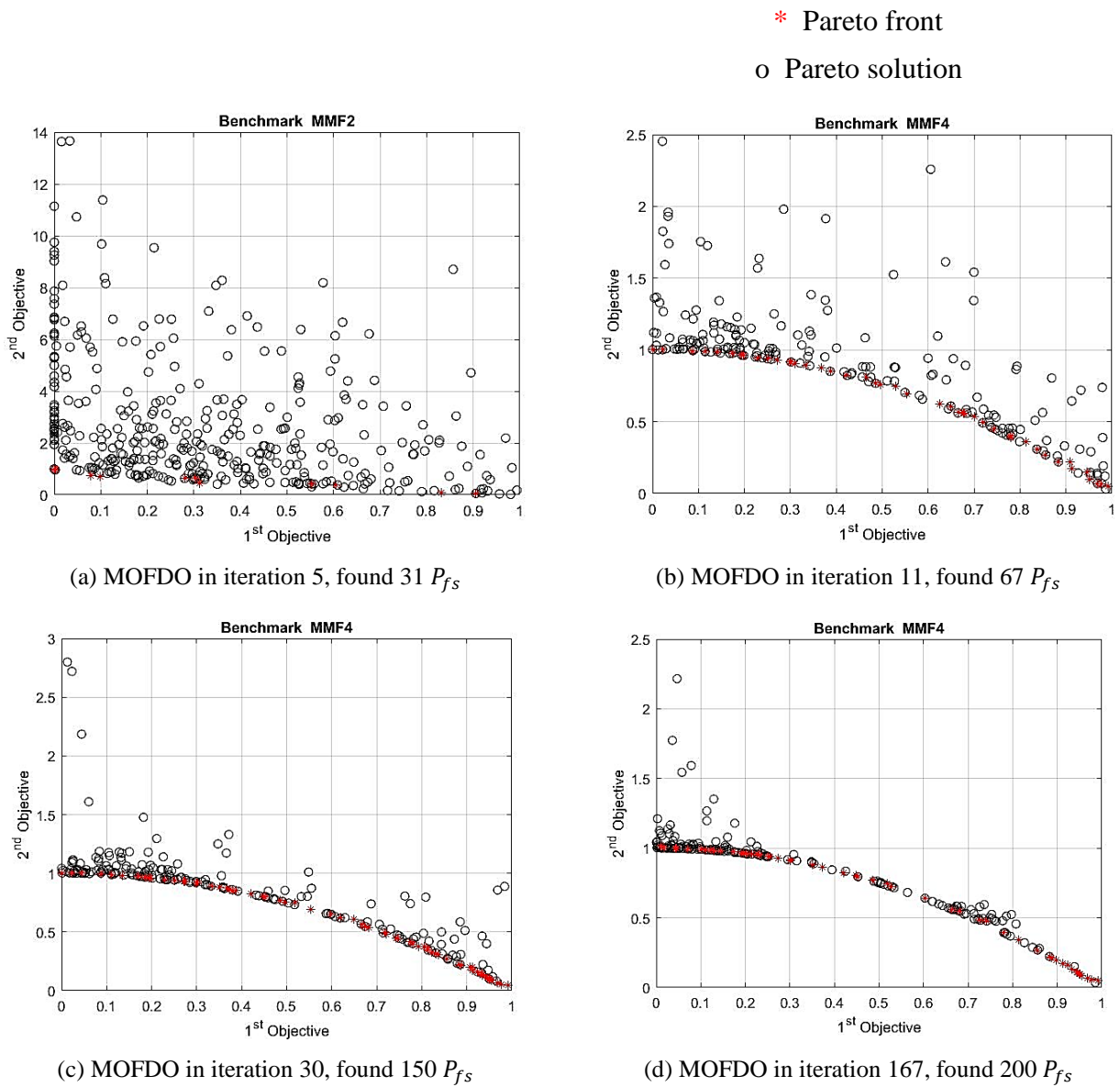

Figure 5. MOFDO solved the MMF4 test function from an initially random solution toward Pareto front optimality.

## 4. Engineering Design Application

MOFDO is implemented in MATLAB codes; the algorithm is well structured, which allows for easy modifications and integration. It also enables researchers to easily understand how it works and how real-world applications can be solved with less effort. To demonstrate this, Welded beam design problem has been optimized with MOFDO, the problem definition and the results are discussed below.

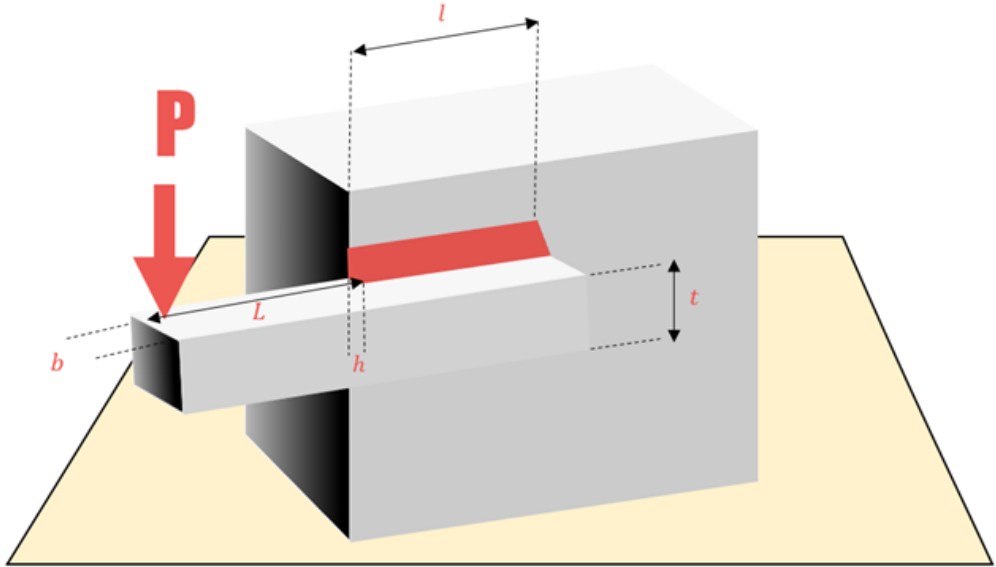

Figure 6. The welded beam design problem

The welded beam design problem is a very well-known real-world engineering design problem, it has been considered by other researchers previously as a test problem for various multi-objective algorithms, such as [58] and [59]. As shown in Figure (6), this problem has four real-parameter variables $x = (h, l, t, b)$, where $h$ is the thickness of the welds, $l$ is the length of the welds, $t$ is the height of the beam, and $b$ is the width of the beam, the $P$ in the Figure (6) represents the amount of load which applies on the beam. This design problem has a bi-objective to be minimized Equation (12), knowing that the objectives are conflicting in nature, the first objective is to minimize the cost of fabrication (X-axis measure in currency) and the second objective is to mitigate the end deflection of the welded beam (Y-axis measure in meters) as follows:

$$\begin{aligned}
& Minimize\ f1(\vec{x}) = 1.10471 h^2 l + 0.04811 tb(14.0 + l), \\
& Minimize\ f2(\vec{x}) = \frac{2.1952}{t^3 b}, \\
& Subject\ to: \\
& \quad g1(\vec{x}) \equiv 13{,}600 - T(\vec{x}) \geq 0, \\
& \quad g2(\vec{x}) \equiv 13{,}600 - \sigma(\vec{x}) \geq 0, \\
& \quad g3(\vec{x}) \equiv b - h \geq 0, \\
& \quad g4(\vec{x}) \equiv Pc(\vec{x}) - 6000 \geq 0, \\
& \quad 0.125 \leq h, \quad b \leq 5.0 \\
& \quad 0.1 \leq l, \quad t \leq 10.0
\end{aligned} \quad (12)$$

As presented in Equation (12), this problem has four constraints to be considered. A violation of any of these constraints will make the design unacceptable. The first constraint is to make sure that the shear stress produced at the beam's support location is less than an allowable value, which is equal to 13,600 psi. The second constraint is to guarantee that the normal stress produced at the beam's support location is less than the acceptable yield strength of the material, which is equal to 30,000 psi. The third constraint is to ensure that the beam's breadth is not less than the weld width from a practical perspective. Constraint number four ensures that the acceptable buckling load $Pc(\vec{x})$ of the beam is larger than the applied load $F = 6000$ lbs. The shear stress $T(\vec{x})$ and the buckling load $\sigma(\vec{x})$ can be calculated as Equation (13 and 14) respectively:

$$T(\vec{x}) = \sqrt{(T')^2 + (T'')^2 + (lT'T'')/\sqrt{0.25(l^2 + (h+t)^2)}}, \qquad (13)$$

$$T' = \frac{6000}{\sqrt{2}\,hl}$$

$$T'' = \frac{6000(14 + 0.5l)\sqrt{0.25(l^2 + (h+t)^2)}}{2\left\{0.707hl(\frac{l^2}{12} + 0.25\,(h+t)^2)\right\}}$$

$$\sigma(\vec{x}) = \frac{504000}{t^2 b}, \qquad (14)$$

$$Pc(\vec{x}) = 64746.022(1 - 0.0282346t)\,tb^3$$

The welded beam design problem is optimized using MOFDO, the algorithm is applied to solve this engineering design problem for 100 iterations, using 100 search agents, and the $P_{fs}$ is stored in 100-sized archives. Figure (7) shows that the obtained $P_{fs}$ are smoothly distributed between these two objectives (Cost and Deflection), and mostly laid on or located very close to the true $P_{fs}$ known in the literature [60]. Also, MOFDO provides a wide variety of feasible solutions for the decision-makers to choose from, this wide variety and smooth distribution of the obtained $P_{fs}$ prove the maturity of the MOFDO in terms of the algorithm capability of tackling real-world engineering design problems effectively.

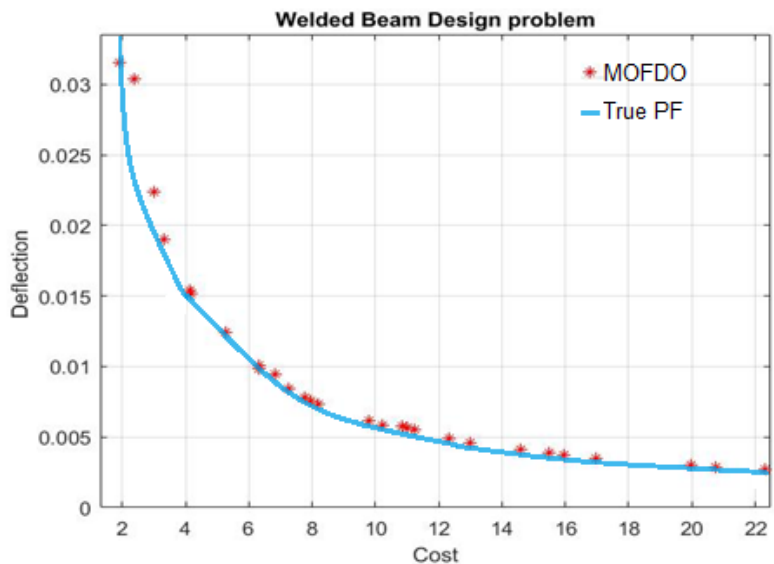

Figure 7. MOFDO results on Welded Beam Design problems

Regarding MOFDO's $P_{fs}$ discovering rate, MOFDO starts from 17 $P_{fs}$ for the first iteration, and then dramatically reaches $100 P_{fs}$ in iteration 36 as presented in Figure (8). This shows how the algorithm efficiently improves multiple initial solutions toward optimality, then occasionally some $P_{fs}$ becomes a non-dominated solution. Hence, they are deleted from the archive; this can be seen from iterations 36 to 100 in Figure (8), the discovery rate of $P_{fs}$ starts from a small number, then steadily increase till it reaches full archive size. This shows that MOFDO is constantly improving all solutions (dominated and non-dominated) throughout all iterations; this feature guarantees that MOFDO avoids local solutions and eventually reaches optimality [61].

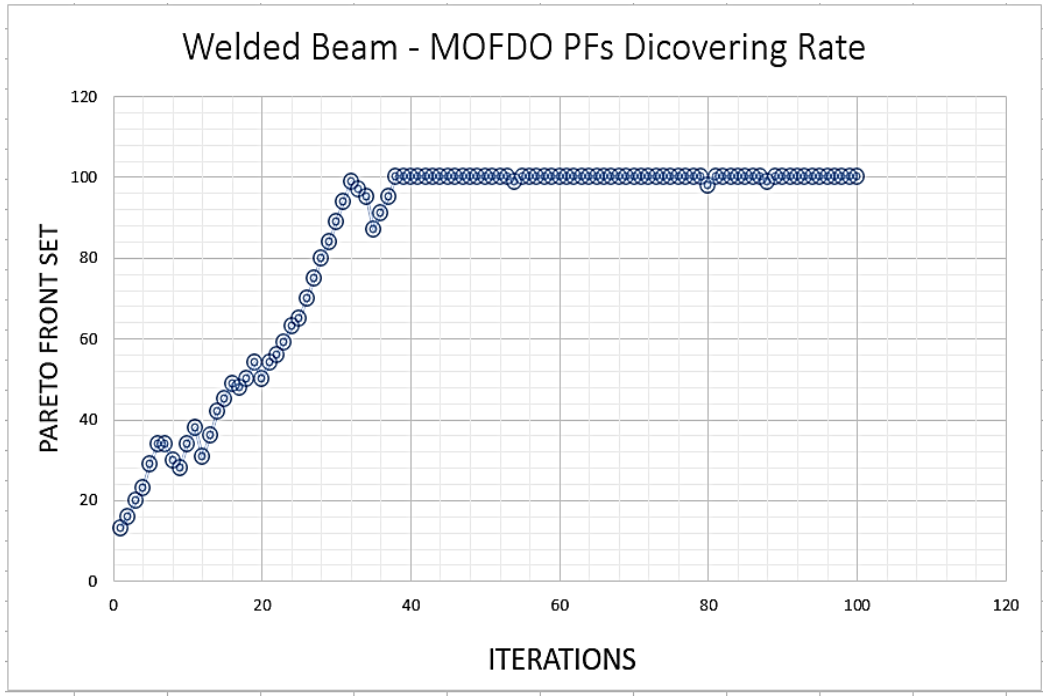

Figure 8. Shows MOFDO Pfs discovering rate

## 5. Conclusions

A multi-objective model for a novel single-objective FDO is proposed, known as a multi-objective fitness dependent optimizer, which is inspired by the bee reproductive swarming process. During the implementation, MOFDO is being treated as a typical cultural algorithm, for this purpose: situational, normative, topographical, domain, and historical knowledge were employed. MOFDO is tested on two different sets of test functions, the first set is ZTD test functions, a classical benchmark used by many other researchers for testing MOEAs. The second set is a modern CEC 2019 multi-modal multi-objective, which is considered to be a more challenging benchmark. The MOFDO results were compared to three other algorithms: the latest state of the art of MOPSO, NSGA-III, and MODA. The comparison showed that MOFDO outperformed other algorithms in most cases and provided comparative results in other cases. MOFDO was easily used for solving real-world problems. For example, the welded beam design problem was solved using MOFDO. It provided well-distributed solutions, which possibly enables decision-makers to have more variant options to consider.

The CEC 2019 benchmark complexity represents a real challenge for MOEAs compared to the classical ZDT benchmark because it contains both local and global Pareto front. The algorithm must try to avoid trapping in the local Pareto front. However, arguably, in some applications, the local optima are preferable, as global optima solutions may not always be applicable in real-world problems [55]. Moreover, having different decision variable space boundary for each dimension make the CEC 2019 MMF even more difficult. The NSGA-III couldn't produce a correct result for the MMF2 test function as presented in Table (2). Also, it has difficulties in some other test functions. On the other hand, MOFDO is constructed to make it easy to deal with these difficulties. Another major drawback of MOPSO is that it is easy to fall into local optimum in high-dimensional space and has a low convergence rate in the iterative process [46], in MOFDO, the $fw$ and $wf$ parameters were used to increase both coverage and convergence of the algorithm, also storing previous good decision for later reuse will convergence speed as well.

Despite that, one of the major contributions of this work is developing MOEA, which has a linear time and space complexity, this means that both time and required space for the algorithm will increase linearly, which is suitable for the current computation architecture. Nonetheless, a polynomial mutation mechanism is employed as a variation operator with the use of an archive for saving the Pareto front solutions. Furthermore, extra storage has been used for saving the previous paces for the potential reuse in future iterations; this will lead to improvement in the algorithm performance Finally, Hypercube grids are used in the implementation to help select the local and global guide individuals. For future works, researchers might try to improve algorithm performance by adapting new parameters, enhancing the learning rate and communication range between individuals, or possibly integrating or hybridizing MOFOD with other MOEAs. Furthermore, there are other interesting multi-objective real-world engineering problems to be optimized by this algorithm, as have been mentioned in the introduction section, such as the four-bar truss design problem, pressure vessel design, the coil compression spring design problem, speed reducer design problem, and car side impact design problem.

**Declaration of interests**

The authors declare that they have no known competing financial interests or personal relationships that could have appeared to influence the work reported in this paper.

The authors declare the following financial interests/personal relationships which may be considered as potential competing interests: None

**Data Availability Statements**

The datasets generated during and/or analysed during the current study are available from the corresponding author upon reasonable request.

[30] D. Brockhoff and E. Zitzler, "mproving Hypervolume-based Multiobjective Evolutionary Algorithms by Using Objective Reduction Methods," in *Congress on Evolutionary Computation (CEC 2007)*, IEEE Press, pp. 2086–2093, 2007.

[31] G. Yu,, "Multi-Objective estimation of Estimation of Distribution Algorithm based on the Simulated binary crossover," *Convergence Information Technology,* vol. 7, no. 13, p. 110–116, 2012.

[32] C.R. Raquel, P.C. Naval,, "An effective use of crowding distance in multiobjective particle swarm optimization," in *the Conference on Genetic and Evolutionary Computation*, Washington, DC, USA, 2005.

[33] S. Mostaghim, J. Teich, , "Strategies for finding good local guides in multi-Objective particle swarm optimization (MOPSO)," in *IEEE Swarm Intelligence Symposium (SIS 2003)*, Indianapolis, IN, USA, 2003.

[34] G.T. Pulido, C. .A. C. Coello,, "Using clustering techniques to improve the performance of a particle swarm optimizer," in *the Genetic and Evolutionary Computation Conference (GECCO)*, Seattle, WA, USA, 2004.

[35] K. D. L. T. E. Zitzler, "Comparison of multiobjective," *Evol Comput,* vol. 8, p. 173–195, 2000.

[36] M. Laumanns, L. Thiele, K. Deb, E. Zitzler, , "Combining Convergence and Diversity in Evolutionary Multiobjective Optimization," *Evolutionary Computation,* vol. 10, no. 3, pp. 263-282, 2002.

[37] L.Tang, X. Wang,, "A Hybrid Multiobjective Evolutionary Algorithm for Multiobjective Optimization Problems," *IEEE Transactions on Evolutionary Computation,* vol. 17, no. 1, pp. 20-45, 2013.

[38] W. Hu, G.G. Yen, X. Zhang,, "Multiobjective particle swarm optimization based on Pareto entropy," *Software,* vol. 25, no. 5, pp. 1025-1050, 2014.

[39] Ganga Negi, Anuj Kumar, Sangeeta Pant, Mangey Ram, "OPTIMIZATION OF COMPLEX SYSTEM RELIABILITY USING HYBRID GREY WOLF OPTIMIZER," *DecisionMaking:Applicationsin Management andEngineering,* vol. 4, no. 2, pp. 241-256, 2021.

[40] M. Das, A. Roy, S. Maity, S. Kar, S. Sengupta, "SOLVING FUZZY DYNAMIC SHIP ROUTING AND SCHEDULING PROBLEM THROUGH MODIFIED GENETIC ALGORITHM," *DecisionMaking: Applicationsin Management andEngineering,* 2021.

[41] R. Stewart and T. S. Palmer, "Utilizing a Reduced-Order Model and Physical Programming," in *PHYSOR2020 – International Conference on Physics of Reactors: Transition to a Scalable Nuclear Future*, 2021.

[42] C. M. Rahman, T. A. Rashid Rashid, A. M. Ahmed and S. Mirjalili, "Multiobjective Learner Performance-based Behavior Algorithm," *Neural Computing and Applications,* vol. 34, no. 8, pp. 6307-6329, 2022.

# 6. Appendix

TABLE 7
ZDT BENCHMARK MATHEMATICAL DEFINITION

| Functions | Mathematical definition |
|---|---|
| ZDT1 | $g(x) = 1 + 9\left(\sum_{i=2}^{n} x_i\right)/(n-1)$<br>$F_1(x) = x_1$<br>$F_2(x) = g(x)\left[1 - \sqrt{x_1/g(x)}\right] \; x \in [0, 1].$ |
| ZDT2 | $g(x) = 1 + 9\left(\sum_{i=2}^{n} x_i\right)/(n-1)$<br>$F_1(x) = x_1$<br>$F_2(x) = g(x)\left[1 - (x_1/g(x))^2\right] \; x \in [0, 1].$ |
| ZDT3 | $g(x) = 1 + 9\left(\sum_{i=2}^{n} x_i\right)/(n-1)$<br>$F_1(x) = x_1$<br>$F_2(x) = g(x)\left[1 - \sqrt{x_1/g(x)} - x_1/g(x)\sin(10\pi x_1)\right] \; x \in [0, 1].$ |

| | | |
|---|---|---|
| ZDT4 | $g(x) = 91 + \sum_{i=2}^{n} \left[ x_i^2 - 10\cos(4\pi x_i) \right]$<br>$F_1(x) = x_1$<br>$F_2(x) = g(x)\left[1 - \sqrt{x_1/g(x)}\right]$ $x_1 \in [0,1], x_i \in [-5,5]$ $i=2,\cdots,10.$ | |
| ZDT5 | $g(x) = 1 + 9[(\sum_{i=2}^{n} x_i)/(n-1)]^{0.25}$<br>$F_1(x) = 1 - \exp(-4x_1)\sin^6(6\pi x_1)$<br>$F_2(x) = g(x)\left[1 - (f_1(x)/g(x))^2\right]$ $x \in [0,1].$ | |

TABLE 8
CEC 2019 MULTI-MODAL MULTI-OBJECTIVE BENCHMARK MATHEMATICAL DEFINITION [55]

| Function | Mathematical definition | Range |
|---|---|---|
| MMF1 | $\begin{cases} f_1 = \|x_1 - 2\| \\ f_2 = 1 - \sqrt{\|x_1 - 2\|} + 2(x_2 - \sin(6\pi\|x_1 - 2\| + \pi))^2 \end{cases}$ | $x_1 \in [1,3],\ x_2 \in [-1,1].$ |
| MMF2 | $\begin{cases} f_1 = x_1 \\ f_2 = \begin{cases} 1 - \sqrt{x_1} + 2(4(x_2 - \sqrt{x_1})^2 \\ \quad -2\cos(\frac{20(x_2 - \sqrt{x_1})\pi}{\sqrt{2}}) + 2),\ 0 \le x_2 \le 1 \\ 1 - \sqrt{x_1} + 2(4(x_2 - 1 - \sqrt{x_1})^2 \\ \quad -\cos(\frac{20(x_2 - 1 - \sqrt{x_1})\pi}{\sqrt{2}}) + 2), 1 < x_2 \le 2 \end{cases} \end{cases}$ | $x_1 \in [0,1],\ x_2 \in [0,2].$ |
| MMF3 | $\begin{cases} f_1 = x_1 \\ f_2 = \begin{cases} 1 - \sqrt{x_1} + 2(4(x_2 - \sqrt{x_1})^2 - \\ 2\cos(\frac{20(x_2 - \sqrt{x_1})\pi}{\sqrt{2}}) + 2) \\ 0 \le x_2 \le 0.5,\ 0.5 < x_2 < 1\ \&\ 0.25 < x_1 \le 1 \\ 1 - \sqrt{x_1} + 2(4(x_2 - 0.5 - \sqrt{x_1})^2 \\ -\cos(\frac{20(x_2 - 0.5 - \sqrt{x_1})\pi}{\sqrt{2}}) + 2) \\ 1 \le x_2 \le 1.5,\ 0 \le x_1 < 0.25\ \&\ 0.5 < x_2 < 1 \end{cases} \end{cases}$ | $x_1 \in [0,1],\ x_2 \in [0,1.5].$ |

| MMF4 | $\begin{cases} f_1 = |x_1| \\ f_2 = \begin{cases} 1 - x_1^2 + 2(x_2 - \sin(\pi|x_1|))^2 & 0 \leq x_2 < 1 \\ 1 - x_1^2 + 2(x_2 - 1 - \sin(\pi|x_1|))^2 & 1 \leq x_2 \leq 2 \end{cases} \end{cases}$ | $x_1 \in [-1, 1], \ x_2 \in [0, 2]$. |
|---|---|---|
| MMF5 | $\begin{cases} f_1 = |x_1 - 2| \\ f_2 = \begin{cases} 1 - \sqrt{|x_1 - 2|} + 2(x_2 - \sin(6\pi|x_1 - 2| + \pi))^2 & -1 \leq x_2 \leq 1 \\ 1 - \sqrt{|x_1 - 2|} + 2(x_2 - 2 - \sin(6\pi|x_1 - 2| + \pi))^2 & 1 < x_2 \leq 3 \end{cases} \end{cases}$ | $x_1 \in [-1, 3], \ x_2 \in [1, 3]$. |
| MMF6 | $\begin{cases} f_1 = |x_1 - 2| \\ f_2 = \begin{cases} 1 - \sqrt{|x_1 - 2|} + 2(x_2 - \sin(6\pi|x_1 - 2| + \pi))^2 & -1 \leq x_2 \leq 1 \\ 1 - \sqrt{|x_1 - 2|} + 2(x_2 - 1 - \sin(6\pi|x_1 - 2| + \pi))^2 & 1 < x_2 \leq 3 \end{cases} \end{cases}$ | $x_1 \in [-1, 3], \ x_2 \in [1, 2]$. |
| MMF7 | $\begin{cases} f_1 = |x_1 - 2| \\ f_2 = 1 - \sqrt{|x_1 - 2|} + \{x_2 - [0.3|x_1 - 2|^2 \cdot \cos(24\pi|x_1 - 2| \\ \qquad + 4\pi) + 0.6|x_1 - 2|] \cdot \sin(6\pi|x_1 - 2| + \pi)\}^2 \end{cases}$ | $x_1 \in [1, 3], \ x_2 \in [-1, 1]$. |
| MMF8 | $\begin{cases} f_1 = \sin|x_1| \\ f_2 = \begin{cases} \sqrt{1 - (\sin|x_1|)^2} + 2(x_2 - \sin|x_1| - |x_1|)^2 & 0 \leq x_2 \leq 4 \\ \sqrt{1 - (\sin|x_1|)^2} + 2(x_2 - 4 - \sin|x_1| - |x_1|)^2 & 4 < x_2 \leq 9 \end{cases} \end{cases}$ | $x_1 \in [-\pi, \pi], \ x_2 \in [0, 9]$. |
| MMF9 | $\begin{cases} f_1 = x_1 \\ f_2 = \dfrac{g(x_2)}{x_1} \end{cases}$ <br> $g(x) = 2 - \sin^6(n_p \pi x)$, $n_p$ is the number of global PSs. | $x_1 \in [0.1, 1.1], \ x_2 \in [0.1, 1.1]$. |
| MMF10 | $\begin{cases} f_1 = x_1 \\ f_2 = \dfrac{g(x_2)}{x_1} \end{cases}$ <br> $g(x) = 2 - \exp\left[-\left(\dfrac{x - 0.2}{0.004}\right)^2\right] - 0.8\exp\left[-\left(\dfrac{x - 0.6}{0.4}\right)^2\right]$. | $x_1 \in [0.1, 1.1], \ x_2 \in [0.1, 1.1]$. |
| MMF11 | $\begin{cases} f_1 = x_1 \\ f_2 = \dfrac{g(x_2)}{x_1} \end{cases}$ | $x_1 \in [0.1, 1.1], \ x_2 \in [0.1, 1.1]$. |
| | $$g(x) = 2 - \exp\left[-2\log(2) \cdot \left(\frac{x-0.1}{0.8}\right)^2\right] \cdot \sin^6(n_p \pi x),$$ $n_p$ is the total number of global and local PSs. | |
| MMF12 | $$\begin{cases} f_1 = x_1 \\ f_2 = g(x_2) \cdot h(f_1, g) \end{cases}$$ $$g(x) = 2 - \exp\left[-2\log(2) \cdot \left(\frac{x-0.1}{0.8}\right)^2\right] \cdot \sin^6(n_p \pi x),$$ $n_p$ is the total number of global and local PSs. $$h(f_1, g) = 1 - \left(\frac{f_1}{g}\right)^2 - \frac{f_1}{g}\sin(2\pi q f_1),$$ $q$ is the number of discontinuous pieces in each PF (PS). | $x_1 \in [0,1],\ x_2 \in [0,1]$. |